\documentclass[conference]{IEEEtran}
\IEEEoverridecommandlockouts
\usepackage{cite}
\usepackage{amsmath,amssymb,amsfonts}
\usepackage{algorithmic}
\usepackage{graphicx}
\usepackage{textcomp}
\usepackage{xcolor}
\usepackage{multicol}
\usepackage{multirow}
\usepackage{graphicx}
\usepackage{url}
\usepackage{hyperref}

\def\BibTeX{{\rm B\kern-.05em{\sc i\kern-.025em b}\kern-.08em
    T\kern-.1667em\lower.7ex\hbox{E}\kern-.125emX}}

\begin{document}

\title{Semantic-guided Representation Learning  for Multi-Label Recognition\\
\thanks{$^{\ast}$Lin Chen and Hezhe Qiao are the corresponding authors. This work is supported by the Chongqing Key Project of Technological Innovation and Application (CSTB2023TIAD-STX0015 and CSTB2023TIAD-STX0031).}
}

\author{\IEEEauthorblockN{\textit{Ruhui Zhang$^{1,2}$, Hezhe Qiao$^{3^\ast}$, Pengcheng Xu$^{1}$, Mingsheng Shang$^{2}$, Lin Chen$^{2^\ast}$} }
\IEEEauthorblockA{
\text{$^{1}$Chongqing University of Posts and Telecommuncation, China} \\
\text{$^{2}$Chongqing Institute of Green and Intelligent Technology, Chinese Academy of Sciences, China} \\
\text{$^{3}$Singapore Management University, Singapore}\\
\text{\{zhangruhui, shangms, chenlin\}} @cigit.ac.cn, hezheqiao@gmail.com, pxu204303@gmail.com}\\
}

\maketitle

\begin{abstract}

Multi-label Recognition (MLR) involves assigning multiple labels to each data instance in an image, offering advantages over single-label classification in complex scenarios. However, it faces the challenge of annotating all relevant categories, often leading to uncertain annotations, such as unseen or incomplete labels. 
Recent  Vision and Language Pre-training (VLP) based methods have made significant progress in tackling zero-shot MLR tasks by leveraging rich vision-language correlations.
However, the correlation between multi-label semantics has not been fully explored, and the learned visual features often lack essential semantic information.
To overcome these limitations, we introduce a Semantic-guided Representation Learning approach (SigRL) that enables the model to learn effective visual and textual representations, thereby improving the downstream alignment of visual images and categories. Specifically, we first introduce a graph-based multi-label correlation module (GMC) to facilitate information exchange between labels, enriching the semantic representation across the multi-label texts. Next, we propose a Semantic Visual Feature Reconstruction module (SVFR) to enhance the semantic information in the visual representation by integrating the learned textual representation during reconstruction. Finally, we optimize the image-text matching capability of the VLP model using both local and global features to achieve zero-shot MLR. Comprehensive experiments are conducted on several MLR benchmarks, encompassing both zero-shot MLR (with unseen labels) and single positive multi-label learning (with limited labels), demonstrating the superior performance of our approach compared to state-of-the-art methods. The code is available at \url{https://github.com/MVL-Lab/SigRL}.

\end{abstract}


\section{Introduction}
\label{sec:intro}

Multi-label recognition (MLR), which assigns multiple labels to each data instance in the image, has found widespread application in areas such as image tagging, document categorization, and medical diagnosis. Despite its critical role and increasing prominence in real-world applications, MLR faces significant challenges in practical scenarios. These challenges include: (1) the exhaustive annotation of image datasets with a large number of categories, which is prohibitively labor-intensive; and (2) the possibility of novel categories (unseen labels) emerging during testing, even with meticulous annotation efforts.

The high cost of labeling can be mitigated by adopting more practical solutions, such as partial-label learning, where training instances are annotated with only a subset of their true labels. To address one of the typical settings, Single Positive Multi-label Learning (SPML), various methods have been proposed, including pseudo-labeling techniques \cite{xie2024class} and novel loss functions \cite{cole2021multi}. However, these approaches face significant limitations in generalizing to unseen labels, particularly when tackling the challenge of recognizing novel categories.


\begin{figure}[!t]
\centering
\includegraphics[width=.9\linewidth]{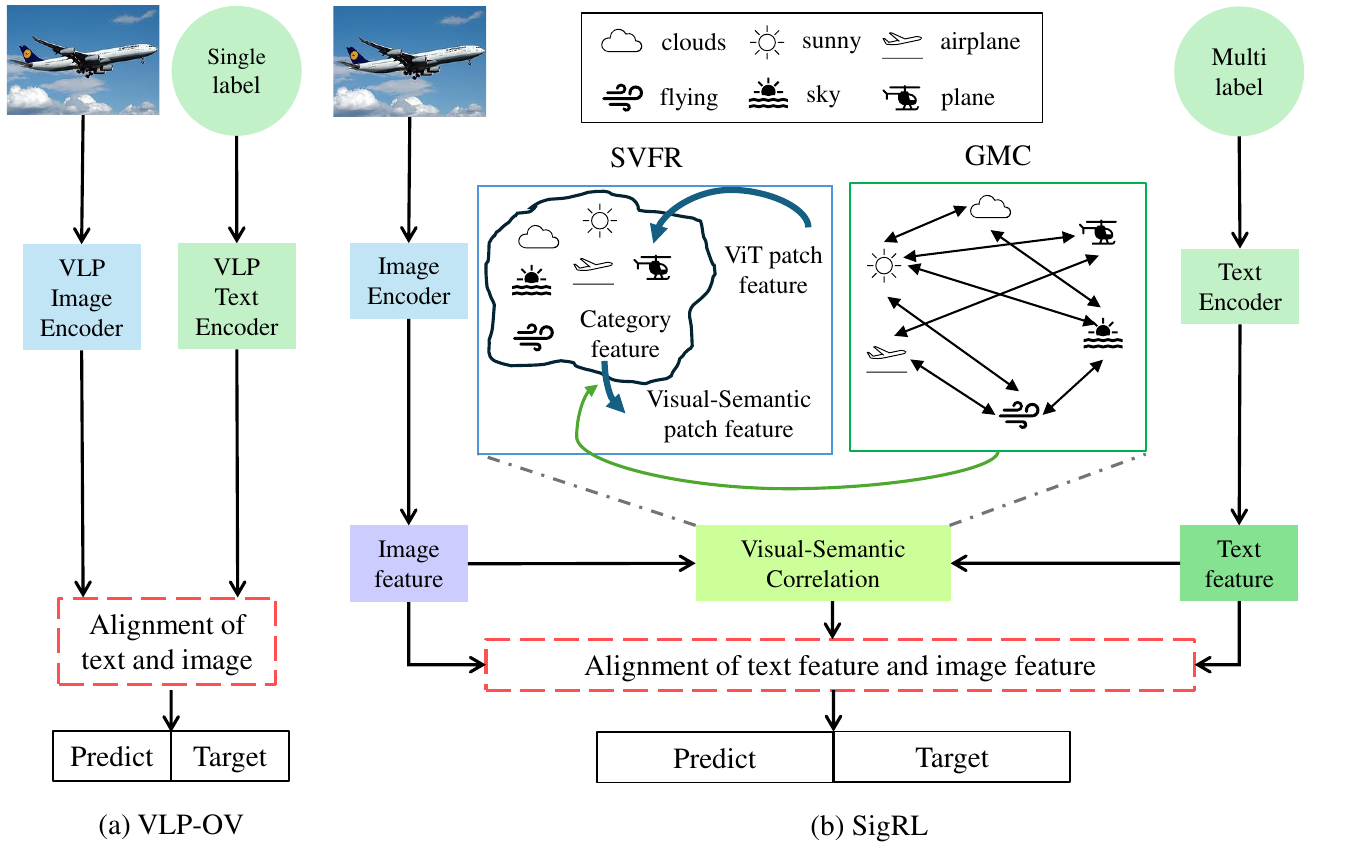}
\caption{
(a) Previous VLP-based methods like VLP-OV \cite{du2022learning} solely focus on modeling and aligning the visual and textual inputs \cite{radford2021learning}.
(b) SigRL leverages GMC to model relationships among multi-label text and uses SVFR to reconstruct semantic visual features, thus improving multimodal representation learning.
}
\label{fig1}
\end{figure}

Recently, Vision and Language Pre-training (VLP) models have demonstrated exceptional performance in multimodal tasks, offering a promising approach for transferring linguistic knowledge from base categories to support open vocabulary (OV) tasks through prompt learning. By pre-training on large-scale image-text pairs \cite{radford2021learning}, VLP models effectively align and fuse cross-modal information by mapping representations of both image and language modalities into a shared feature space. 
However, most VLP-OV models  \cite{du2022learning} primarily focus on single-label classification and have received limited attention in the context of MLR with unseen labels, also known as zero-shot MLR, as illustrated in Fig.~\ref{fig1} (a). To address this issue, He et al.\cite{he2023open} proposed the multi-modal knowledge (MKT) approach for zero-shot MLR, using knowledge distillation to align patch features with multi-labels. Xing et al. \cite{xing2024vision} leverage CLIP \cite{radford2021learning} to generate robust positive and negative pseudo-labels for zero-shot MLR. 

However, these approaches fail to thoroughly explore the correlation between multi-labels, and the learned visual features lack essential semantic information. While some studies \cite{ye2020attention} utilize Graph Neural Networks (GNNs) with predefined, static graph structures to model inter-label dependencies, these methods are not specifically tailored for zero-shot MLR, resulting in suboptimal performance.

Therefore, we propose a semantic-guided representation learning approach (SigRL). SigRL explores the relationships between multiple labels and enriches semantic information by integrating text features into visual representations during reconstruction, as shown in Fig.~\ref{fig1}(b). This general framework explicitly models the interaction between semantic image features and multiple labels, enhancing text-image alignment within VLP models.

Our contributions can be summarized as follows:

\begin{itemize}
    \item We propose a Semantic-guided Representation Learning approach (SigRL) to effectively learn the representations of visual images and text, which can benefit the downstream alignment between visual images and categories.
    
    \item We further introduce a graph-based Multi-label correlation (GMC) and Semantic Visual Feature Reconstruction (SVFR) to implement SigRL. GMC constructs a graph structure to capture the intricate semantic relationships among labels, while SVFR integrates semantic features into the visual representation during reconstruction to learn effective visual representations.

    \item Extensive experiments on three widely used large-scale datasets demonstrate that our proposed SigRL achieves state-of-the-art performance on both zero-shot MLR and SPML tasks.
    
\end{itemize}

\section{Method}
\subsection{Problem Definition}
In this study, we primarily focus on two types of MLR with uncertain annotations: zero-shot MLR and SPML tasks. The training dataset consists of $N$ samples, $\mathcal{X}=\left\{ \textbf{X}_k, {\textbf{Y}_k} \right\} _{k=1}^{N} $
where $\textbf{X}_k$ denotes the $k$-th image and ${\textbf{Y}_k}$ is a set of labels containing $C$ classes. Each label $y_k^i \in \textbf{Y}_k, i \in [1,C]$ can be 1 (positive), -1 (negative), or 0 (unknown).

In zero-shot MLR, the dataset is divided into two classes: base classes $\textbf{Y}^B$ and novel classes $\textbf{Y}^U$. $\textbf{Y}_k\in \textbf{Y}^B$ represents the known label for the training image, while $\textbf{Y}^U$ denotes the set of unseen labels that are not present in the training data.
For VLP model, such as CLIP, which has a strong vocabulary $\textbf{L}$, this vocabulary does not strictly cover $\textbf{Y}^B$ and $\textbf{Y}^U$, and may also include predefined words outside of the new categories. Model training is not solely focused on $\textbf{Y}^B$ but rather on $\textbf{L}$.
Zero-shot learning (ZSL) and generalized zero-shot learning (GZSL) are two primary tasks within zero-shot MLR. 
ZSL focuses on learning a classifier $f_{zsl}$ to identify the unseen categories as: ${f_{zsl}}:\mathcal{X} \to {\textbf{Y}^U}$.
GZSL extends this by requiring the model to classify both seen and unseen categories, \textit{i.e.}, ${f_{gzsl}}:\mathcal{X} \to {\textbf{Y}^B\cup \textbf{Y}^U}$.

In SPML task, ${\textbf{Y}_k}$ associated with an image contains only one positive label. Specifically, for each $y_k^j$ in $\textbf{Y}_k$, $y_k^j \in \{0,1 \} ^c$ and $\sum\nolimits_{j = 1}^C {{\mathbb{I}_{\left[ {y_k^j = 1} \right]}}}=1$, where $\mathbb{I_{\left[.\right]}}$ denotes the indicator function, signifying that only one label in $\textbf{Y}_k$ is positive.
Although the training data is highly sparse, the test task requires predicting multiple labels. In this work, we randomly select and retain a single positive label for each training image, treating all other labels as unknown. This selection is performed once per dataset. The validation and test sets remain fully labeled.

\begin{figure*}[ht]
\centering
\includegraphics[width=.9\linewidth]{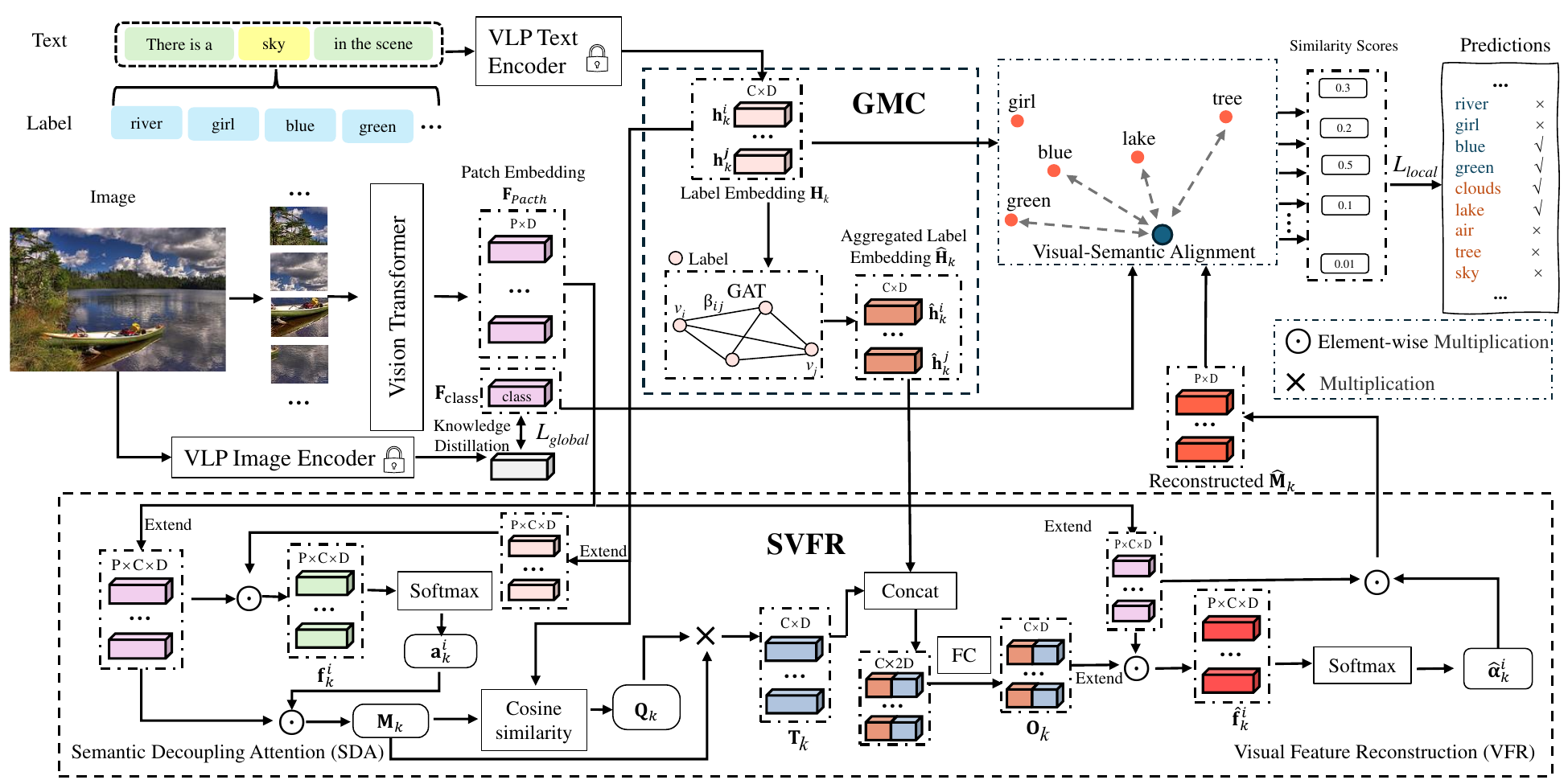}
\caption{ The overview of SigRL. 
The VLP Image Encoder and VLP Text Encoder are employed to extract features from the image and label, respectively. The proposed GMC is then utilized to construct a fully connected graph.
The SVFR consists of two components: SDA and VFR. SDA decouples visual features and integrates semantic information, while VFR reconstructs visual features by aligning spatial information encoded within ViT patch features. 
}  

\label{fig3}
\end{figure*}

\subsection{The Overall Framework} 

The overall structure of the SigRL model is illustrated in Fig.~\ref {fig3}. SigRL comprises two main modules: a Graph-based Multi-label Correlation module (GMC) and a Visual Feature Reconstruction module (VFR). Specifically, a VLP image encoder is employed to extract features from both images and constructed text. The patch features are obtained by distilling knowledge from the VLP encoder. In the GMC, SigRL captures label relationships by constructing a topology where labels can exchange information. Meanwhile, the VFR integrates the corresponding semantic information into the visual representation by reconstructing the patch embeddings. Finally, SigRL aligns the visual image and labels based on a similarity score.

\subsection{Graph-based Multi-label Correlation Modeling}
The pre-trained VLP model CoOp \cite{zhou2022learning} serves as the VLP text encoder and the corresponding label embedding ${\textbf{h}^i}$ of the image is obtained by
\begin{equation}
\label{gt}
{\textbf{h}^i}=\text{CoOp}\left( {\textbf{Y}_i} \right).
\end{equation} 

To exploit the potential relations among labels, we build a fully connected graph for each image where each node represents the label and the attribute of each node is the learned embedding $\textbf{h}^i$. Then we applied the Graph Attention Network (GAT) \cite{brody2021attentive} to dynamically adjust the influence of neighboring labels through an attention mechanism in information exchange. GAT assign a score $e\left( v_{i},v_{j} \right)$ for the edge which connects node $i$ and 
node $j$, indicating the importance of the characteristics of the r node $v_{j}$ to the node $v_{j}$. 

\begin{equation}
\label{eq4}
e\left( v_{i},v_{j} \right) =\text{LeakyReLU}\left( \alpha  ^T \left[ \textbf{W}\textbf{h}^{i},\textbf{W}\textbf{h}^{j} \right] \right),
\end{equation}
where $v_j \in {\cal N}({v_i})$ indicates the neighboring node of node $v_{i}$ in the graph. $\textbf{W}$ denotes the trainable matrix mapping the attribute to the latent space, and $\alpha $ is the learnable attention vector.  Then the normalization is performed on the attention score $e\left( v_{i},v_{j} \right)$ to obtain the attention coefficient by:
\begin{equation}
\label{eq5}
\beta_{i j}=\frac{\exp \left(e{(v_i, v_j)}\right)}{\sum_{v_k \in \mathcal{N}(v_i)} \exp \left(e(v_i, v_k)\right)}.
\end{equation}

Finally, the normalized attention coefficient $\beta_{i j}$ is used to compute the new representation of each node.
\begin{equation}
\label{eq6}
\mathbf{\hat{h}}^{i}=\sigma \left( \sum_{v_j\in {\cal N}({v_i})}{\beta_{ij} \textbf{W}\textbf{h}^{j}} \right),
\end{equation}
where $\sigma$ denotes the nonlinear activate function. 
The aggregated label embedding ${\hat{\textbf{H}}_k}=[\hat{\textbf{h}}^1_k,\dots,\hat{\textbf{h}}^C_k]$ 
with rich semantic information will be integrated into the visual representation, where $C$ is the number of classes.

\subsection{Semantic Visual Feature Reconstruction} 

To bridge the semantic gap given that the learned visual features often lack essential semantic information, we propose the SVFR, which leverages a multi-label semantic separation strategy to learn category-specific visual representations, thereby benefiting the downstream alignment of visual images and categories. 


We denote the patches as $\textbf{V}_k$, $\textbf{V}_k=\left[ \textbf{F}_{class},\textbf{F}_{patch} \right]$, where  $\textbf{F}_{class}$ and $\textbf{F}_{patch}$ represent the outputs of the class and patch tokens, respectively.  $\textbf{F}_{patch} \in \mathbb{R}^{P \times D}$, where $P$ is the number of patches, $D$ represents the output dimension of the Vision Transformer(ViT) \cite{dosovitskiy2020image}. Concurrently, from the VLP text encoder, we can obtain label embeddings $\textbf{H}_{k} \in \mathbb{R}^{C\times D}$ for the $k$-th image by text prompt, where each row $\textbf{h}_{k}^{i} \in \mathbb{R}^{1\times D}$ represents the embedding for category $i$, along with the same dimension $D$ as the patch features.

The SVFR framework comprises two key components: semantic decoupling attention and visual feature reconstruction. The semantic decoupling attention module disentangles the features extracted by the ViT by projecting them onto distinct category-specific subspaces, enabling it to effectively capture the interaction between visual and semantic information.

\noindent \textbf{Semantic Decoupling Attention (SDA).}
In SDA, we first extend the dimensions of both $\textbf{F}_{patch}$ and $\textbf{H}_{k}$ to $\mathbb{R}^{P\times C \times D}$. Then the semantic-guided attention mechanism is employed to adaptively weight the contribution of different spatial and semantic features using the attention score $\textbf{f}_{k}^{i}$ which is formulated as  
\begin{equation}
\label{eq7}
\textbf{f}_{k}^{i}=FC(\text{tanh}(\textbf{F}_{patch}\odot \textbf{h}_{k}^{i})),
\end{equation}
where $FC(\cdot)$ is a fully connected layer , and $\odot$ denotes the element-wise multiplication operation.  The category-specific features $\textbf{M}_{k}$ is defined as the weighted sum of image features $\textbf{f}_{k}^{i}$,
\begin{equation}
\label{eq9}
\textbf{M}_k=\sum\limits_{i = 1}^C{\textbf{a}_{k}^{i}\textbf{F}_{patch}},
\end{equation}
where the attention coefficients $\textbf{a}_{k}^{i}$ is derived from $\textbf{h}_{k}^{i}$ and $\textbf{F}_{patch}$, which is formulated as the following
 
\begin{equation}
\label{eq8}
\textbf{a}_{k}^{i}=softmax(FC(\text{tanh}(\textbf{f}_{k}^{i})).
\end{equation}

In addition, the category-specific attention map $\textbf{q}_{k}^{ij} \in \mathbf{Q}_{k}^{i}$ representing the correlation between the label embedding $\mathbf{h}_{k}^{j}$  and the acquired category-specific features $\mathbf{m}_{k}^{i} \in \mathbf{M}_{k}$ is calculated with cosine similarity by: 
\begin{equation}
\label{eq10}
\textbf{q}_{k}^{i,j}=RELU\left( \frac{\left( \mathbf{m}_{k}^{i} \right) ^T\mathbf{h}_{k}^{j}}{\lVert \mathbf{m}_{k}^{i} \rVert \lVert \mathbf{h}_{k}^{j} \rVert} \right).
\end{equation}

Then the normalization is performed on the visual feature vectors of each category to obtain $\textbf{T}_{k}^{i}$  representing semantic decoupled features is defined as the following
\begin{equation}
\label{eq11}
\textbf{T}_{k}^{i}=Norm\left( \textbf{Q}_{k}^{i} \right) \times \textbf{M}_{k}^{i}.
\end{equation}


By decoupling the patch feature based on the label embedding, the SigRL model integrates the original label information through the above operations to enrich the semantic information in the visual feature.

\noindent \textbf{Visual Feature Reconstruction (VFR).} Directly projecting them into $\mathbb{R}^{C \times D}$ risks the loss of this crucial spatial context by mapping the dimensions of the patch's representation into the dimension of class representation. To mitigate the loss of information, we further propose the VFR to reconstruct the representation of visual images by aligning the spatial information encoded in ViT patch features.

We first concatenate $\mathbf{\hat{H}}_{k}$ and $\textbf{T}_{k}$ and then transformed it through a fully connected layer to yield $\textbf{O}_{k}\in \mathbb{R}^{C \times D}$. 
Then, $\textbf{O}_{k}\in \mathbb{R}^{C \times D}$ is concatenated with the image $\textbf{F}_{patch}$ to generate semantic visual features as:
  
\begin{equation}
\label{eq12}
\mathbf{\hat{f}}_{k}^{i}=FC\left( \text{tanh}\left( \textbf{F}_{patch}\odot \textbf{O}_{k}^{i} \right) \right) 
\end{equation}

Then, the vectors $\textbf{O}_{k}$ and $\textbf{F}_{patch}$ are processed with an attentional function (\textit{i.e.}, a fully connected operation). Subsequently, a softmax function is applied to normalize the resulting values and obtain the attention coefficients $\mathbf{\hat{a}}_{k}^{i}$ as follows
\begin{equation}
\label{eq13}
\hat{\textbf{a}}_{k}^{i}=softmax\left( \mathbf{\hat{f}}_{k}^{i} \right).
\end{equation}
Finally, the reconstructed representation with rich semantic information  $\hat{\textbf{M}}_{k} \in \mathbb{R}^{P \times D}$ associated with the patches are obtained using the weighted average pooling of patches, which is formulated as the following:

\begin{equation}
\label{eq14}
\hat{\textbf{M}}_{k}=\sum_{P}^{}{\mathbf{\hat{a}}_{k}^{i}\textbf{F}_{patch}}.
\end{equation}

\subsection{Alignment of Visual Image and Category} 
\label{loss}
By fusing the semantic and visual features, the predicted scores $s_i$ for each category are computed based on the $\textbf{F}_{patch} = \left\{ \hat{\textbf{M}}_{k}^{1},\hat{\textbf{M}}_{k}^{2},\cdots ,\hat{\textbf{M}}_{k}^{P} \right\}$ and $\textbf{F}_{class}$, which is formulated as
\begin{equation}
\label{eq16}
\resizebox{\hsize}{!}{
${s_i=\left< \textbf{h}^{i},\textbf{F}_{class} \right> +\text{Top-K}\left( \left\{ \left< \textbf{h}^{i}\cdot \hat{\textbf{M}}_{k}^{1} \right> ,\cdot \cdot \cdot ,\left< \textbf{h}^{i}\cdot \hat{\textbf{M}}_{k}^{P} \right> \right\} \right)},$}
\end{equation}
where $\textbf{h}^{i} \in \mathbb{R}^{1 \times D}$ denotes the original label embedding for each category, $\left< \cdot\right>$  denotes inner product, and Top-K ($\cdot$) refers to $top-k$ mean pooling. 

To align the visual image and category, we employ a total loss function $L_{total}$ consisting of  $L_{global}$ and $L_{local}$
\begin{equation}{
L_{total}=L_{local}+L_{global}},
\end{equation}
where the $L_{local}$ is defined based on a ranking loss to enhance the representational capacity of local features extracted by the ViT,
\begin{equation}
\label{eq17}
L_{local}=\sum_k^{}{\sum_{p\in \textbf{Y}^i_k,n\notin \textbf{Y}^i_k}^{}{max \left( 1+s_{k}^{n}-s_{k}^{p},0 \right)}},
\end{equation}
where $\textbf{Y}^i_k \in \textbf{Y}^B$ denotes the ground truth labels of base classes for $k$-th image, and $s_{k}^{n}$ and $s_{k}^{p}$ represent the scores of negative label $n$ and positive label $p$, respectively.
$L_{global}$ is a knowledge distillation loss function defined based on the $\textbf{F}_{class}$ as the following 
\begin{equation}
\label{eq15}
L_{global}=\lVert \text{CLIP}\left( \textbf{X}_{k} \right) -\textbf{F}_{class} \rVert _{1},
\end{equation}
where $\textbf{X}_{k}$ denotes the feature of $k$-th image, CLIP($\cdot$) denotes the output of the pre-trained CLIP image encoder, and $\lVert \cdot \rVert {_1}$ represents the ${L_1}$ loss, known as the mean absolute error.


\section{Experiments}
\subsection{Experiments Setup}
\noindent
\textbf{Datasets.}
We evaluate the proposed SigRL on three widely used datasets.  The NUS-WIDE dataset \cite{chua2009nus}, containing 925 labels, is employed for the zero-shot MLR  task. In this setting, 925 labels are considered seen, while the remaining 81 labels are unseen. The Microsoft COCO (COCO) \cite{lin2014microsoft} includes 80 object categories, with 82,801 training and 40,504 validation images. VOC (VOC)\cite{everingham2010pascal} covers 20 categories, offering 5,011 training/validation and 4,952 test images. Following previous studies \cite{he2023open, sun2022dualcoop}, we evaluated models using mean average precision (mAP), F1, Picison (P), and Recall (R) for the Top-K predictions.

\noindent
\textbf{Implementation Details.}
The ViT-B/16 model pre-trained on ImageNet-1K is used as the ViT backbone of the SigRL model.
For each image with the resolution of $224\times 224$, ViT-B/16 splits it into $16\times 16$ patches.
Regarding hyperparameter settings, $k$ in $top-k$ pooling is set to 16 defined in the SigRL, and the ViT feature dimension $D$ is set to 512.
The optimizer employs AdamW with a base learning rate of 1e-4, a minimum learning rate of 1e-7, and a weight decay of 0.05. 

\begin{table*}[ht]
\begin{center}
\caption{State-of-the-art comparison for ZSL and GZSL tasks on the NUS-WIDE dataset. The top-performing score is highlighted in bold, while the second-best score is underlined.} 
\scalebox{0.93}{
\label{tab1}
\begin{tabular}{| c | c  c  c | c  c  c | c | c  c  c | c  c  c | c |}
\hline
\multirow{3}{*}{Methods} & \multicolumn{7}{c|}{Zero-Shot Learning (ZSL)} & \multicolumn{7}{c|}{Generalized zero-Shot Learning (GZSL)}\\
& \multicolumn{3}{c|}{Top-3} & \multicolumn{3}{c|}{Top-5} & \multirow{2}{*}{mAP} & \multicolumn{3}{c|}{Top-3} & \multicolumn{3}{c|}{Top-5} & \multirow{2}{*}{mAP}\\  
& P & R & F1 & P & R & F1 & & P & R & F1 & P & R & F1 &\\
\hline

LabelEM (2015)\cite{akata2015label} & 15.60 & 25.00 & 19.20 & 13.40 & 35.70 & 19.50 & 7.10 & 15.50 & 6.80 & 9.50 & 13.40 & 9.80 & 11.30 & 2.20\\
Fast0Tag (2016)\cite{zhang2016fast} & 22.60 & 36.20 & 27.80 & 18.20 & 48.40 & 26.40 & 15.10 & 18.80 & 8.30 & 11.50 & 15.90 & 11.70 & 13.50 & 3.70\\
OAL (2018)\cite{kim2018bilinear} & 20.90 & 33.50 & 25.80 & 16.20 & 43.20 & 23.60 & 10.40 & 17.90 & 7.90 & 10.90 & 15.60 & 11.50 & 13.20 & 3.70\\
$\text{LESA}_{\text{M}10}$(2020)\cite{huynh2020shared} & 25.70 & 41.10 & 31.60 & 19.70 & 52.50 & 28.70 & 19.40 & 23.60 & 10.40 & 14.40 & 19.80 & 14.60 & 16.80 & 5.60\\
BiAM (2021)\cite{narayan2021discriminative} & - & - & 33.10 & - & - & 30.70 & 26.30 & - & - & 16.10 & - & - & 19.00 & 9.30\\
$\text{SDL}_{\text{M}7}$(2021\cite{ben2021semantic} & 24.20 & 41.30 & 30.50 & 18.80 & 53.40 & 27.80 & 25.90 & 27.70 & \underline{13.90} & 18.50 & 23.00 & \underline{19.30} & 22.10 & 12.00\\
CLIP-FT (2021)\cite{radford2021learning} & - & - & 23.50 & - & - & 21.70 & 30.50 & - & - & 20.30 & - & - & 23.20 & 16.80\\
MKT(2023)\cite{he2023open} & - & - & 34.10 & - & - & 31.10 & \underline{37.60} & - & - & \underline{22.00} & - & - & \underline{25.40} & \underline{18.30}\\
Gen-MLZSL(2023)\cite{gupta2023generative}& \underline{26.60} & \underline{42.80} & 32.80 & \underline{20.10} & \underline{53.60} & 29.30 & 25.70 & \underline{30.90} & 13.60 & 18.90 & \underline{26.00} & 19.10 & 22.00 & 8.90 \\
$\left( \text{ML} \right) ^2\text{P}\_\text{Encoder}$(2023)\cite{liu20232} & - & - & 32.80 & - & - & $\textbf{32.30}$ & 29.40 & - & - & 16.80 & - & - & 19.20 & 10.20 \\
SCP(2024)\cite{ma2024label} & - & - & \underline{33.80} & - & - & 31.00 & 27.20 & - & - & 16.10 & - & - & 18.80 & 10.10 \\
SigRL(Ours) & $\textbf{27.73}$ & $\textbf{44.32}$ & $\textbf{34.11}$ & $\textbf{21.77}$ & $\textbf{58.01}$ & \underline{31.66} & $\textbf{38.46}$ & $\textbf{39.43}$ & $\textbf{17.38}$ & $\textbf{24.13}$ & $\textbf{32.60}$ & $\textbf{23.95}$ & $\textbf{27.61}$ & $\textbf{20.85}$\\
\hline 
\end{tabular}
}
\end{center}
\end{table*} 

\subsection{Main Results}
In the experiments, we performed two separate tasks: zero-shot MLR and SPML. 
In the zero-shot MLR comparison, we compared the SigRL model with several state-of-the-art methods published in recent years. This comparison included a diverse set of techniques, including Vision-Language Pre-training (VLP)-based models, \textit{e.g.}, CLIP model with ranking loss (CLIP-FT), MKT, Gen-MLZSL models\cite{gupta2023generative}, $\left( \text{ML} \right) ^2\text{P}\_\text{Encoder}$\cite{liu20232} and SCP\cite{ma2024label}.


The experimental results for zero-shot MLR are summarized in Table \ref{tab1}. Notably, the VLP model CLIP-FT achieves significantly higher performance compared to traditional ZSL approaches. While MKT achieved the second-highest F1 and mAP scores on GZSL by focusing on multi-modal alignment between labels and image features. Our SigRL further leverages multi-label information for image feature reconstruction, leading to enhanced performance in zero-shot MLR. Quantitatively, SigRL outperformed MKT by 2.13 and 2.21 in Top-3 and Top-5 F1 scores for GZSL, respectively, and also achieved state-of-the-art results on the ZSL task.
Meanwhile, the number of parameters in SigRL is 59.82M, which is only 4.47\% more than the base model MKT.

\begin{figure}[!t]
\centering
\resizebox{0.4\textwidth}{!}{\includegraphics{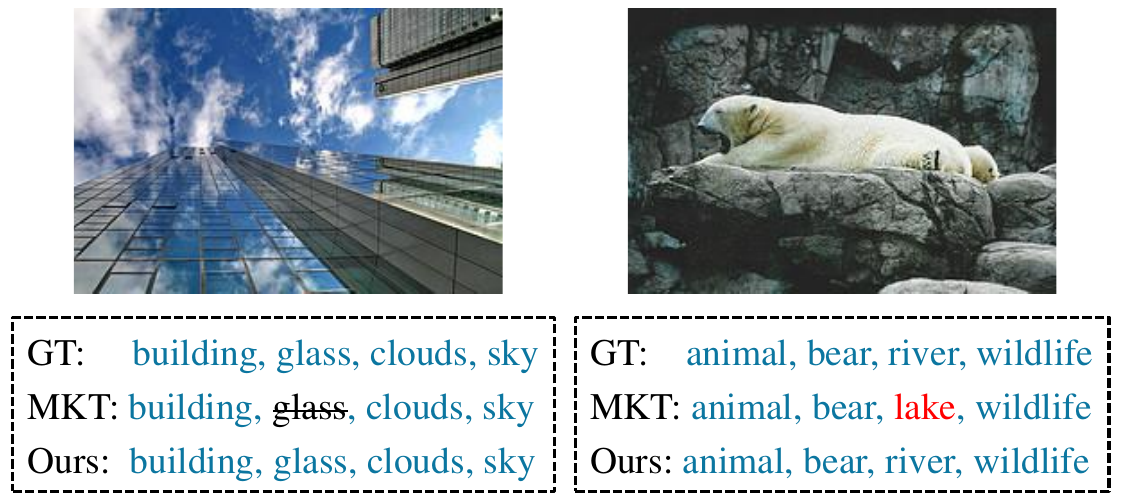}}
\caption{
The qualitative analysis of zero-shot MLR, where ground truth labels are denoted by GT, correct predictions are shown in blue, incorrect predictions are shown in red, and missed predictions are denoted by strikeout.}

\label{fig4}
\end{figure}

Figure \ref{fig4} illustrates zero-shot MLR predictions on NUS-WIDE for the tested models. Our proposed model accurately predicted both "building" and "glass", as well as correctly inferred the relationship between "bear" and "river", thereby correctly predicting "lake". In contrast, MKT failed to predict "glass" and misclassified "river" as "lake", highlighting the importance of multi-label semantic visual features for enhanced VLP performance.

\begin{table}[!t]
\begin{center}
\caption{Comparison for SPML tasks on the COCO dataset.}
\label{tab2}
\resizebox{\columnwidth}{!}{
\begin{tabular}{| c | c | c | c | c  c  c | c  c  c |}
\hline
\multirow{2}{*}{Ann. Labels} & \multirow{2}{*}{Methods} & \multirow{2}{*}{Loss} & \multirow{2}{*}{mAP} & \multicolumn{3}{c|}{Top-3} & \multicolumn{3}{c|}{Top-5}\\
&  &  &  & P & R & F1 & P & R & F1 \\
\hline
\multirow{7}{*}{1 P.\&0N.} & DualCoOp\cite{sun2022dualcoop} & - & 69.20 & - & - & - & - & - & - \\
\cline{2-10}
& \multirow{3}{*}{MKT\cite{he2023open}} & {IUN}\cite{cole2021multi} & 73.65 & 63.92 & 76.99 & 69.85 & 45.56 & 85.67 & 59.49\\
& & {EM\_APL}\cite{zhou2022acknowledging} & 74.66 & \underline{65.85} & \underline{78.92} & \underline{71.80} & \underline{46.62} & 87.32 & \underline{60.79}\\
& & {EM}\cite{zhou2022acknowledging} & 73.72 & 65.06 & 78.05 & 70.97 & 46.16 & 86.65 & 60.23\\
\cline{2-10}
& \multirow{3}{*}{SigRL(Ours)} & {IUN}\cite{cole2021multi} & 74.19 & 65.43 & 77.64 & 70.48 & 45.82 & 86.07 & 59.80\\
& & {EM\_APL}\cite{zhou2022acknowledging} & \underline{74.70} & 65.63 & 78.83 & 71.63 & 46.58 & \underline{87.41} & 60.77\\
& & {EM}\cite{zhou2022acknowledging} & $\textbf{74.86}$ & $\textbf{65.87}$ & $\textbf{78.99}$ & $\textbf{71.84}$ & $\textbf{46.68}$ & $\textbf{87.45}$ & $\textbf{60.87}$\\

\hline
\end{tabular}}
\end{center}
\end{table} 

\begin{table}[!t]
\begin{center}
\caption{Comparison for SPML tasks on the VOC dataset.}
\label{tab3}
\resizebox{\columnwidth}{!}{
\begin{tabular}{| c | c | c | c | c  c  c | c  c  c |}
\hline
\multirow{2}{*}{Ann. Labels} & \multirow{2}{*}{Methods} & \multirow{2}{*}{Loss} & \multirow{2}{*}{mAP} & \multicolumn{3}{c|}{Top-3} & \multicolumn{3}{c|}{Top-5}\\
&  &  &  & P & R & F1 & P & R & F1 \\
\hline
\multirow{7}{*}{1 P.\&0N.} & {DualCoOp}\cite{sun2022dualcoop} & - & 83.60 & - & - & - & - & - & - \\
\cline{2-10}
& \multirow{3}{*}{MKT\cite{he2023open}} & {IUN}\cite{cole2021multi} & \underline{86.57} & 43.25 & 93.86 & 59.47 & 27.30 & 96.57 & 42.57\\
& & {EM\_APL}\cite{zhou2022acknowledging} & 84.58 & \underline{44.23} & \underline{94.53} & \underline{60.26} & \underline{27.76} & \underline{97.45} & \underline{43.21}\\
& & {EM}\cite{zhou2022acknowledging} & 86.25 & 43.31 & 92.94 & 59.09 & 27.32 & 96.26 & 42.56\\
\cline{2-10}
& \multirow{3}{*}{SigRL(Ours)} & {IUN}\cite{cole2021multi} & $\textbf{87.20}$ & 43.74 & 94.03 & 59.71 & 27.46 & 96.87 & 42.79 \\
& & {EM\_APL}\cite{zhou2022acknowledging} & 84.60 & 43.11 & 93.04 & 58.92 & 27.35 & 96.54 & 42.62\\
& & {EM}\cite{zhou2022acknowledging} & 86.45 & $\textbf{44.50}$ & $\textbf{95.28}$ & $\textbf{60.67}$ & $\textbf{27.90}$ & $\textbf{98.01}$ & $\textbf{43.44}$\\
\hline
\end{tabular}}
\end{center}
\end{table} 
To evaluate the performance of SigRL on SPML tasks, we compared it with other models using different SPML loss functions, including Ignores Unobserved Negatives (IUN) loss \cite{cole2021multi}, entropy-maximization (EM) loss \cite{zhou2022acknowledging}, and entropy-maximization asymmetric pseudo-labeling (EM\_APL) \cite{zhou2022acknowledging}. The results on the COCO and VOC datasets are presented in Tables \ref{tab2} and \ref{tab3}, respectively. We employ three loss functions: ignores unobserved negatives (IUN) \cite{cole2021multi} loss, entropy-maximization (EM) \cite{zhou2022acknowledging} loss for SPML task, and entropy-maximization asymmetric pseudo-labeling (EM\_APL) \cite{zhou2022acknowledging} loss, which were compared across different models.
Notably, the experimental results of DualCoOp \cite{sun2022dualcoop}, were reported by Xing et al. \cite{xing2024vision}. Consistent with the original DualCoOp \cite{sun2022dualcoop}, we employ ResNet-101 as the visual encoder, resizing images to 448x448 pixels.
Table \ref{tab2} presents the experimental results on the COCO dataset. The EM loss function yields the most significant improvement. Compared to MKT, our model achieves a 1.14 increase in mAP and 0.87 and 0.64 improvements in Top-3 and Top-5 F1 scores, respectively.
Table \ref{tab3} demonstrates the effectiveness of our model on the VOC dataset. Compared to MKT, which also uses the IUN loss function, our model achieves a 0.20 higher mAP and 1.88 and 0.88 higher Top-3 and Top-5 F1 scores, respectively. 
The observed performance gains can be attributed to SigRL's mechanism of reconstructing features with richer semantic and image information, thereby enhancing multi-modal alignment by leveraging the properties of the VLP model, even with only limited valid annotations.

\begin{table*}[!t]
\begin{center}
\caption{Ablation experiment for ZSL and GZSL tasks on the NUS-WIDE.}
\label{tab4}
\begin{tabular}{| c  c  c | c  c  c | c  c  c | c | c  c  c | c  c  c | c |}
\hline
\multicolumn{3}{|c|}{Methods} & \multicolumn{7}{c|}{Zero-Shot Learning (ZSL)} & \multicolumn{7}{c|}{Generalized zero-Shot Learning (GZSL)}\\
&  &  & \multicolumn{3}{c|}{Top-3} & \multicolumn{3}{c|}{Top-5} & \multirow{2}{*}{mAP} & \multicolumn{3}{c|}{Top-3} & \multicolumn{3}{c|}{Top-5} & \multirow{2}{*}{mAP}\\
{GMC} & {SDA} & {VFR} & P & R & F1 & P & R & F1 & & P & R & F1 & P & R & F1 &\\
\hline
$\times$ & $\times$ & $\times$& 25.67 & 41.04 & 31.59 & 19.92 & 53.08 & 28.97 & 37.60 &  37.33 & 16.46 & 22.85 & 30.89 & 22.70 & 26.17 & 19.98\\
\checkmark & $\times$ & $\times$ & 25.73 & 41.13 & 31.66 & 20.28 & 54.04 & 29.49 & \underline{38.44} & 38.13 & 16.81 & 23.33 & 31.74 & 23.32 & 26.89 & 20.06\\
\checkmark & \checkmark & $\times$  & \underline{26.57} & \underline{42.48} & \underline{32.69} & \underline{20.77} & \underline{55.35} & \underline{30.21} & 38.09 &  \underline{39.05} & \underline{17.22} & \underline{23.90} & \underline{32.39} & \underline{23.80} & \underline{27.44} & \underline{20.62}\\
\checkmark & \checkmark & \checkmark & $\textbf{27.73}$ & $\textbf{44.32}$ & $\textbf{34.11}$ & $\textbf{21.77}$ & $\textbf{58.01}$ & $\textbf{31.66}$ & $\textbf{38.46}$ & $\textbf{39.43}$ & $\textbf{17.38}$ & $\textbf{24.13}$ & $\textbf{32.60}$ & $\textbf{23.95}$ & $\textbf{27.61}$ & $\textbf{20.85}$\\
\hline 
\end{tabular}
\end{center}
\end{table*}

\subsection{Ablation Study}
To evaluate the contribution of each module, we conducted ablation studies on the SigRL model. As shown in Table \ref{tab4}, the experimental results demonstrate that these modules improve the VLP model's performance on zero-shot MLR. Notably, the VFR module within the SigRL model contributed to a 0.37 improvement in mAP for the ZSL task. 
For GZSL, the GMC module led to a 0.08 improvement in mAP and 0.48 and 0.72 improvements in F1 for Top-3 and Top-5, respectively. Furthermore, the SVFR, which consists of the SDA and VFR modules, improved performance across all test metrics. The enhanced MLR performance in SVFR is attributed to the fusion of semantic and visual features, and the GMC's capacity to model multi-label correlations.



\section{Conclusion}

In this paper, we propose a novel semantic guide representation learning approach SigRL, which provides a unified framework for improving MLR under two scenarios including zero-shot MLR and SPML. In SigRL,  we first introduce the GMC to exploit semantic relationships between multiple labels by incorporating a trainable graph learning structure during the representation learning of text. Second, to enrich the semantic information of the representation of visual images, we further introduce the SVFR to integrate the semantic information into the representation of visual images during the reconstruction. By doing this SigRL can learn effective representation of visual images which enhance the alignment of text-image pairs in the VLP model.  Finally, the alignment of visual images and categories is achieved by optimizing the knowledge distill loss and ranking loss from global and local features respectively.  We conducted experiments on the zero-shot MLR task using the NUS-WIDE dataset and the SPML task using the COCO and VOC datasets. The experimental results demonstrate the significant effectiveness of our method.


\bibliographystyle{IEEEbib}
\bibliography{reference.bib}

\vspace{12pt}

\end{document}